\documentclass{article}

\usepackage[preprint,nonatbib]{neurips_2022}
\usepackage[utf8]{inputenc} 
\usepackage[T1]{fontenc}    
\usepackage{hyperref}       
\usepackage{url}            
\usepackage{booktabs}       
\usepackage{amsfonts}       
\usepackage{amsmath}        
\usepackage{bm} 
\usepackage{nicefrac}       
\usepackage{microtype}      
\usepackage{xcolor}         
\usepackage{graphicx}   

\title{Approximate sampling and estimation of partition functions using neural networks}

\author{%
  George T.~Cantwell\\
  Santa Fe Institute, 1399 Hyde Park Road, Santa Fe, NM, 87501\\
  \texttt{gcant@umich.edu} \\
}

\begin{document}

\maketitle

\begin{abstract}
We consider the closely related problems of sampling from a distribution known up to a normalizing constant, and estimating said normalizing constant.  We show how variational autoencoders (VAEs) can be applied to this task.  In their standard applications, VAEs are trained to fit data drawn from an unknown and intractable distribution.  
We invert the logic and train the VAE to fit a simple and tractable distribution, on the assumption of a complex and intractable latent distribution, specified up to normalization.  This procedure constructs approximations without the use of training data or Markov chain Monte Carlo sampling. We illustrate our method on three examples: the Ising model, graph clustering, and ranking.
\end{abstract}

\section{Background}

Many problems that arise in statistics, combinatorics, and physics can be understood through the lens of an appropriate probability distribution, together with its normalizing constant.
To study objects $\bm{x}$, we consider the distribution $P(\bm{x}) = f(\bm{x}) / \mathcal{Z}$, where $\mathcal{Z} = \sum_{\bm{x}} f(\bm{x})$.

For example, when considering a combinatorial problem we can set $f(\bm{x})=1$ if a set of constraints is satisfied and $f(\bm{x})=0$ otherwise.
Then, $\mathcal{Z}$ simply counts the total number of solutions to the constraints.
In a physics setting, we could have $f(\bm{x}) = e^{-\beta H(\bm{x})}$, where $H(\bm{x})$ is an energy function and $\beta$ the inverse temperature.  The resulting distribution is known as the canonical distribution, and arises when a system is in thermal equilibrium with its surroundings.
The quantity $\mathcal{Z}$ is called the \emph{partition function}, and encodes the thermodynamic properties of the system \cite{baxter2016exactly}.
In a statistical inference problem we can set $f(\bm{x}) = P(\bm{x}, \mathcal{D})$, where $\mathcal{D}$ is the observed data.  The resulting distribution $f(\bm{x}) / \mathcal{Z}$ is the posterior distribution for $\bm{x}$. The quantity $\mathcal{Z} = P(\mathcal{D})$ is known as the \emph{model evidence} or \emph{marginal likelihood} and is important for rigorous model selection \cite{mackay2003information}.
Understanding such distributions, and gaining a handle on $\mathcal{Z}$, is a perennial problem across computer science, statistics, and physics.

Computing $\mathcal{Z}$ exactly is generally difficult -- even for many of the simplest examples the problem is \#P-hard \cite{moore_nature_2011}.
The naive approach is to enumerate all exponentially many $\bm{x}$, and directly sum over this set; \#P-hardness means there is unlikely to be any substantially faster method than this.
But while we usually cannot compute $\mathcal{Z}$ exactly, it may still be possible to make useful approximations.

A standard approach for approximating $\mathcal{Z} = \sum_{\bm{x}} f(\bm{x})$ is to use Markov chain Monte Carlo (MCMC) \cite{levin2017markov, newman1999monte, landau2021guide}.
This requires constructing a Markov chain that has $P(\bm{x}) = f(\bm{x}) / \mathcal{Z}$ as its equilibrium distribution, and then simulating the process.
While finding such a Markov chain can be straightforward, it often turns out that the process takes exponential time to approach its equilibrium distribution.
If the chain takes exponential time to get close to the equilibrium distribution, it is ultimately little improvement over the naive but exact method.
Even when Markov chains do mix in polynomial time, they may still scale with a large power of the problem size, and thus large problems are out of reach in practice.

The contribution of this paper is a new method for approximating $\mathcal{Z}$ and sampling $\bm{x} \sim f(\bm{x}) / \mathcal{Z}$, using neural networks.
This provides an alternative to MCMC for practical problems in Bayesian statistics and physics.

The neural network architecture we employ is equivalent to a variational autoencoder (VAE) \cite{kingma2013auto, kingma2019introduction}.
An autoencoder is a neural network that is trained to map objects to themselves, i.e., to represent the identity function.
To make this non-trivial, constraints are added to the intermediate representations.
An autoencoder is constructed from two parts, known as the encoder and the decoder.
In a \emph{variational} autoencoder, the encoder maps complicated data, such as photographs of human faces, to a simple latent distribution, such as a (multivariate) normal or Bernoulli distribution.
The decoder takes points from the latent space and attempts to reconstruct a posterior distribution for the data (e.g. a distribution over faces).
The basic setup takes data~$\rightarrow$ latent distribution~$\rightarrow$ reconstructed data distribution, where the input data should match the output distribution.

We will essentially reverse the logic of VAEs.
As discussed, in the standard application of VAE we train the network to map samples from an intractable data distribution to a simple latent distribution, and then back again.
Here, in contrast, we will train the VAE to map a simple and tractable distribution to a more complex one.
The only input to the training procedure will be a routine to compute $f(\bm{x})$.
From this function alone (i.e., without any samples to be used as training data), the VAE will attempt to find a good approximation for $\mathcal{Z} = \sum_{\bm{x}} f(\bm{x})$, and will construct a sampler for $\bm{x} \sim f(\bm{x})/\mathcal{Z}$.

At a conceptual level, this procedure is similar to normalizing flows: we take a simple distribution and learn a mapping to a more interesting one \cite{9089305}.
Because normalizing flows use the Jacobian change-of-variables transformation, they require every step in the neural network to be a bijection (invertible) and are not readily applied to discrete distributions.
To circumvent this issue, Ref.~\cite{altosaar2020probabilistic} advocates training two networks with normalizing flows.
The approach is conceptually close to ours although we do not use normalizing flows, and so our networks are considerably less constrained.
In particular, the dimensions of the inputs and outputs in our framework do not need to match, which allows us to fit low entropy but high dimensional distributions by using high entropy but low dimensional inputs -- indeed, this is the key insight that makes VAEs useful.
Further, by adapting the framework of autoencoders, all of the flexibility and advances in this field can be incorporated into the approximations.
Finally, there is also a conceptual similarity between our goals and those in \cite{PhysRevLett.122.080602}, which uses autoregressive networks.
Both methods attempt to fit distributions, although again, ours is both more simple and more flexible, since it does not require autoregressive constraints, and the networks can thus be much smaller.
In fact, as we shall see, our method can perform well in certain regimes with only a single binary input/output neuron (in which case, optimization can actually be done with a pen and paper).

In the remaining sections of this paper we first derive the required machinery, and then test it against three example problems.

\section{Variational approximations and field theories}

Variational methods turn the problem of estimating $\mathcal{Z}$ into an optimization problem.
We note that
\begin{equation}
	\ln \mathcal{Z} = \ln \sum_{\bm{x}} f(\bm{x}) \geq \sum_{\bm{x}} R(\bm{x}) \ln \frac{f(\bm{x})}{R(\bm{x})}
	\label{eq:jensen1}
\end{equation}
where $R(\bm{x})$ is any arbitrary distribution.
This inequality is a special case of Jensen's inequality.

While the inequality in Eq.~(\ref{eq:jensen1}) is true for all $R(\bm{x})$,
maximizing the right hand side with respect to $R$ provides the best lower bound for $\ln \mathcal{Z}$.
In fact, the unconstrained optimum is simply $R(\bm{x})=P(\bm{x})=f(\bm{x})/\mathcal{Z}$, in which case the inequality in Eq.~(\ref{eq:jensen1}) is saturated and $\ln \mathcal{Z}$ is computed exactly.
Of course, this optimum is not practically useful, since by assumption $P(\bm{x})$ is difficult to work with.
Instead we must restrict $R(\bm{x})$ to some sufficiently simple family of distributions, such as product distributions $R(\bm{x}) = \prod_i r_i(x_i)$.

In physics, this approach is known as a mean-field approximation \cite{mackay2003information}.
If the $x_i \in \{\pm1\}$ then we can write
\begin{equation}
	R(\bm{x}) = \prod_i { e^{x_i \phi_i} \over e^{\phi_i} + e^{-\phi_i} }
\end{equation}
and $\phi_i$ is conceptualized as a field felt by $x_i$, characterizing the average (mean) effect of the rest of the system.
Mean-field approximations are often effective but can be greatly inaccurate.
Unfortunately, improving the approximation is often quite involved.

Moving beyond mean-field theory, we can allow the fields to fluctuate and develop what is known as a statistical field theory \cite{brezin2010introduction}.
Equivalently, $R$ is a mixture of product distributions, rather than a simple product distribution.
Constructing such approximations is somewhat of an art, and often involves highly problem-specific insights (e.g. the Hubbard-Stratonovich transformation for the Ising model \cite{PhysRevLett.3.77}).

Due to the complex and technical nature of developing statistical field theories, the approach is not nearly as widely applied as mean-field theory.
For statistics problems in particular, where the precise details of the model usually change with each application, developing a statistical field theory is usually unreasonably difficult. 
Here we discuss a partial remedy to this.

\subsection{Auxiliary variables, statistical fields}

To make progress in our goal of constructing an improved approximation for $P(\bm{x})=f(\bm{x})/\mathcal{Z}$, we introduce auxiliary variables $\bm{y}$. 
We couple the distributions of the variables $\bm{x}$ and $\bm{y}$ as
\begin{equation}
	P(\bm{x}, \bm{y}) = P(\bm{x}) Q( \bm{y} \vert \bm{x} )
	\label{eq:joint_distribution}
\end{equation}
where $Q(\bm{y} \vert \bm{x})$ is an arbitrary normalized distribution, and hence $\sum_{\bm{y}} Q(\bm{y} \vert \bm{x})=1$.
Using the fact that $Q$ is normalized we now invoke Jensen's inequality to find
\begin{equation}
	\ln \mathcal{Z} =  \ln \sum_{ \bm{x}, \bm{y}} f(\bm{x}) Q(\bm{y}\vert\bm{x})  \geq \sum_{ \bm{x}, \bm{y}} P(\bm{y}) R(\bm{x}\vert\bm{y}) \ln \frac{  f(\bm{x}) Q(\bm{y}\vert\bm{x})}{P(\bm{y})R(\bm{x}\vert \bm{y})}
	\label{eq:joint_Jensen}
\end{equation}
for all distributions $P(\bm{y})$, $Q(\bm{x} \vert \bm{y})$, and $R(\bm{x}\vert\bm{y})$. 
Now, a double maximization of Eq.~(\ref{eq:joint_Jensen}) with respect to both $Q(\bm{y} \vert \bm{x})$ and $R(\bm{x}\vert \bm{y})$  provides a lower bound for $\ln \mathcal{Z}$.

Equation~(\ref{eq:joint_Jensen}) provides a framework to estimate $\ln \mathcal{Z}$. 
In addition, it also suggests a method to sample $\bm{x}\sim P(\bm{x})$.
The $Q$ and $R$ that maximize Eq.~(\ref{eq:joint_Jensen}) satisfy
\begin{equation}
	P(\bm{x}) Q(\bm{y}\vert\bm{x}) = P(\bm{y}) R(\bm{x}\vert\bm{y}).
\end{equation}
This means that if we were to sample $\bm{y}$ from $P(\bm{y})$, and then $\bm{x}$ from $R(\bm{x}\vert\bm{y})$, the $\bm{x}$ we end up with is distributed according to $P(\bm{x})$.
The problem, of course, is to find good $Q$ and $R$.

As was the case for Eq.~(\ref{eq:jensen1}), the true optimum that saturates the inequality in Eq.~(\ref{eq:joint_Jensen}) is probably useless in practice.
Rather, we must restrict ourselves to cases where $Q$ and $R$ are tractable, such as a product of conditional distributions
\begin{equation}
	 R( \bm{x} \vert \bm{y} ) = \prod_i r_i(x_i \vert \bm{y}),
\end{equation}
and we let $P(\bm{y})$ be a simple and tractable distribution, for example a normal or Bernoulli distribution.

Returning to the case where $x_i \in \{ \pm 1 \}$ we can, without loss of generality, write the product as 
\begin{equation}
	R(\bm{x} \vert \bm{y}) = \prod_i {e^{x_i \phi_i(\bm{y})} \over  e^{\phi_i(\bm{y})} + e^{-\phi_i(\bm{y})}}.
	\label{eq:conditional_product_R}
\end{equation}
The functions $\phi_i(\bm{y})$ now act as fields, although they fluctuate because $\bm{y}$ is a random variable; this is a statistical field theory.
We shall represent $\bm{\phi}(\bm{y})$ using a neural network.
The expectation of Eq.~(\ref{eq:conditional_product_R}) over the auxiliary variables $\bm{y}$ provides  an approximation for the true $P(\bm{x})$.  
How we proceed now depends on our choice of distribution $P(\bm{y})$.

A convenient choice for $P(\bm{y})$ is that each $y_i$ is an independent coin flip: $+1$ or $-1$ with probability $1/2$.
In this case we can use the same ansatz for $Q$ as we did for $R$, Eq.~(\ref{eq:conditional_product_R}),
\begin{equation}
	Q(\bm{y} \vert \bm{x}) = \prod_i {e^{y_i \theta_i(\bm{x})} \over  e^{\theta_i(\bm{x})} + e^{-\theta_i(\bm{x})}}.
\end{equation}
The function $\bm{\theta}(\bm{x})$ can be represented in a neural network.

Alternatively, we could choose $P(\bm{y})$ to be a product of independent normal distributions, in which case the posterior $Q(\bm{y} \vert \bm{x})$ will (hopefully) be approximately normal about some mean $\bm{\mu}(\bm{x})$ with covariance matrix $\bm{\Sigma}(\bm{x})$.
Specifically, we assume $Q( \bm{y} \vert \bm{x} ) = \mathcal{N}_{ \bm{\mu}(\bm{x}), \bm{\Sigma}(\bm{x}) } ( \bm{y})$, with $\mathcal{N}$ being the density function for the normal distribution.  Optimal choices for $\bm{\mu}$ and $\bm{\Sigma}$ are
\begin{align}
	\bm{\mu}(\bm{x}) = \operatorname*{arg\,max}_{\bm{y}} R(\bm{x} \vert \bm{y}) \quad\quad \text{and}\quad\quad 
	\bm{\Sigma}^{-1}(\bm{x}) = - \left. {\partial^2 \ln R(\bm{x} \vert \bm{y}) \over \partial y_i \partial y_j }\right\vert_{\bm{y}=\bm{\mu}(\bm{x})},
	\label{eq:Laplace_approx}
\end{align}
which is equivalent to the Laplace approximation \cite{mackay2003information,gelman2013bayesian} of Bayesian statistics.
Two problems arise: (i)~the $\operatorname*{arg\,max}_{\bm{y}}$ operation is expensive and (ii)~we do not have easy access to gradient information.
To fix these problems we can abandon Eqs.~(\ref{eq:Laplace_approx}) and instead train a second neural network to represent $\bm{\mu}(\bm{x})$ and $\bm{\Sigma}(\bm{x})$.

Putting everything together, we maximize Eq.~(\ref{eq:joint_Jensen}) using stochastic optimization.
To do this we draw $\bm{y}$ from $P(\bm{y})$ and then $\bm{x}$ from $R(\bm{x} \vert \bm{y})$.
When $\bm{x}$ is discrete, as it often will be, we can apply the Gumbel softmax trick  \cite{jang2016categorical, maddison2016concrete} to ensure we can take derivatives through the sampling process.
Each term in Eq.~(\ref{eq:joint_Jensen}) is now simple to compute, and so are its derivatives, and we have a complete system.

We simultaneously train two networks.
The first network represents $R(\bm{x}\vert\bm{y})$ and transforms each specific value of the auxiliary variables $\bm{y}$ into a distribution over the variables of interest, $\bm{x}$.
The second network represents $Q(\bm{y}\vert\bm{x})$ and does the reverse: it takes specific values of $\bm{x}$ and maps them to a distribution over auxiliary variables $\bm{y}$.
As discussed this is equivalent to a VAE, except the role of the encoder and decoder have been conceptually reversed; unlike the standard application, we do not train the neural networks on samples $\bm{x}$ from the intractable distribution.
In fact, the networks are trained to match the distribution of $\bm{x}$ without ever seeing a sample of $\bm{x}$.

\section{Examples}

\subsection{Square lattice Ising model}

The Ising model is one of the most widely studied models from physics, and has been applied well beyond physics \cite{baxter2016exactly, stauffer2008social}.
Originally formulated as a model for magnetism, each site (i.e. atom) has a small magnetic dipole caused by ``spin'' that either points up or down.
Due to an exchange interaction (a quantum mechanical effect) it is energetically favorable for  neighboring spins to align.
The spin at site $i$ is denoted $x_i \in \{ \pm 1 \}$ and the energy function is
\begin{equation}
	H(\bm{x}) = - J \sum_{ \langle i,j \rangle} x_i x_j - h \sum_i x_i
\end{equation}
where $J$ is the coupling strength, $\langle i,j \rangle$ represent neighboring pairs, and $h$ is an external magnetic field.
When a majority of the spins are aligned, we observe a magnetic moment at macroscopic scales.

The model has been studied on a myriad of topologies, but the archetypal case is the two-dimensional square lattice.
When $h=0$ the square lattice Ising model has be solved exactly, both at finite size and in the limit of an infinite system \cite{baxter2016exactly, moore_nature_2011}.
For $h\neq 0$ it remains an open problem.

So that we can compare our results to exact calculations we will set $h=0$ and $J=1$, in which case we have
\begin{equation}
	f(\bm{x}) = e^{\beta \sum_{ \langle i,j \rangle} x_i x_j}.
\end{equation}
Where $\langle i,j \rangle$ represent neighboring pairs on the square lattice, and we wrap around at the boundaries so the the left-most sites are connected to the right-most, and likewise for the top and bottom.

We consider a system of $n=64$ spins, and use simple architectures for the neural networks.
Specifically, both neural networks have only a single hidden layer with $1024$ units using the SELU activation function, and for the Gumbel softmax function, ${e^{w_i/\tau} \over \sum_j e^{w_j/\tau}}$, we set $\tau=1/16$.
For all experiments we used the Adam optimizer with the default parameters in TensorFlow $2.8$, and trained on a desktop computer with an Intel i5-7600K CPU, and no dedicated GPU.

For the auxiliary variables $\bm{y}$ we used independent Bernoulli variables.
It is not immediately obvious how many auxiliary variables are needed, and so we experimented with differing numbers.
By Eq.~(\ref{eq:joint_Jensen}), any number of auxiliary variables provides a bound; we experimented with different numbers (between $1$ and $64$) and took whichever estimate of $\ln \mathcal{Z}$ was largest.

\begin{figure}
  \centering
  \includegraphics[width=1.0\linewidth]{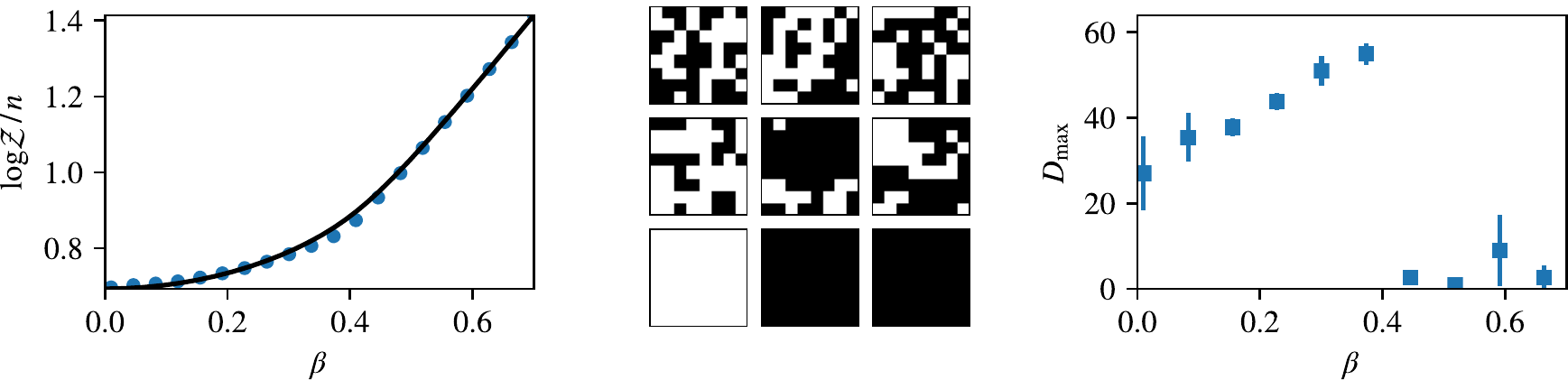}
	\caption{\label{fig:Ising} Experiments for the square lattice Ising model on $n$ nodes.  Left: estimates of ${1\over n}\ln \mathcal{Z}$ for different values of $\beta$.  The solid line shows the analytic result. Center: Random samples from the trained network, at three different temperatures, from top to bottom: $\beta = 0.01, 0.35, 0.7$.  Qualitatively, these samples match expectations.  At $\beta=0.01$ we see a mostly random pattern, at $\beta=0.35$ we see larger contiguous blocks, and at $\beta=0.7$ we see a homogeneous system. Right: Networks were trained with differing sized input distributions, i.e. different numbers of auxiliary variables.  Since the resulting approximation of $\ln \mathcal{Z}$ is a rigorous lower bound, regardless of the number of auxiliary variables, we ask which value of $D$ -- the number of auxiliary variables -- gives the largest lower bound.  For $\beta$ larger than the critical value, the best approximations have a small number of auxiliary variables.  
	}
\end{figure}

The results of our experiments are presented in Fig.~\ref{fig:Ising}.
Once training is finished, we have estimates of the partition function along with an (approximate) sampler for the model.
To generate states we first draw $\bm{y}$ from $P(\bm{y})$ and then $\bm{x}$ from $R(\bm{x}\vert \bm{y})$.
Each new sample is independent and generated in time proportional to the size of the neural network.
We can use these samples to probe arbitrary properties of the model.

The standard method for sampling from the Ising model is to use Markov chain Monte Carlo methods \cite{newman1999monte, landau2021guide}.
Considerable effort has been spent to develop efficient Markov chains, although mixing times are still generally worse than linear in the system size.

We should also contrast the approach presented here to some previous studies of the Ising model using autoencoders, which also have the ability to approximately sample \cite{yevick2022variational, walker2020deep, d2020learning}.
The key difference is that in these works, the Ising model is first simulated using Markov chain Monte Carlo methods, and then the results of the simulations fit using an autoencoder.
The resulting fit is, of course, limited by the quality of the initial samples.
In other words, this approach requires that one already has a high-quality Monte Carlo sampler, whereas in our approach no input data are required -- we fit the distribution directly.

\subsection{Graph clustering and the Stochastic Block Model}

The Stochastic Block Model (SBM) is a generative statistical model for the graph clustering problem, also known community detection \cite{holland1983109, FORTUNATO20161, moore2017computer}.

The model generates networks on $n$ nodes as follows.
First, each node is randomly assigned to a community.
Let $x_i$ be the community assignment for node $i$.
Once each node is assigned to a community, edges are created independently at random.
An edge between node $i$ and $j$ exists with probability $\omega({x_i, x_j})$ -- a symmetric function of the community membership of nodes $i$ and $j$ only.

The probability to generate graph $G$ is 
\begin{equation}
	P(G \vert \bm{x}, \omega) = \prod_{i<j} \omega(x_i, x_j)^{G_{ij}} ( 1 - \omega(x_i,x_j) )^{1-G_{ij}} 
	\label{eq:SBM1}
\end{equation}
where $G_{ij}$ is an indicator variable for the existence of edge $(i,j)$.

To invert the generative model, and hence to find a good community assignment $\bm{x}$ for a given graph $G$, we could simply maximize Eq.~(\ref{eq:SBM1}) with respect to both $\bm{x}$ and $\omega$.
Each value $x_i$ can be thought of as a parameter, and we are then finding the maximum likelihood community assignment.
However, because there are $n$ community parameters, i.e. the number of parameters is growing linearly with the size of the network, this approach is liable to over-fitting.
A principled approach to inference is to instead maximize $P(G \vert \omega) = \sum_{\bm{x}} P(G, \bm{x} \vert \omega)$.

Assuming a uniform prior on $\bm{x}$ over $c$ different groups we have the posterior
\begin{equation}
	P(\bm{x} \vert G, \omega) = \frac{c^{-n} \prod_{i<j} \omega(x_i, x_j)^{G_{ij}} ( 1 - \omega(x_i,x_j) )^{1-G_{ij}}}{\mathcal{Z(\omega)} }
	\label{eq:SBM_posterior}
\end{equation}
and where $\mathcal{Z(\omega)} = P(G \vert \omega)$.

For the purpose of an example we will consider the case of two groups -- $a$ and $b$ -- with connection probabilities $\omega(a,a) = \omega(b,b) = \omega_{\mathrm{in}}$ and $\omega(a,b) = \omega_{\mathrm{out}}$.
We take $f(\bm{x})$ to be the numerator of Eq.~(\ref{eq:SBM_posterior}), and we consider the standard benchmark graph -- the Karate Club graph \cite{zachary1977information}.

The neural network architectures are unchanged from the previous example -- $1024$ units in a single hidden layer, with SELU activations, and again we set $\tau=1/16$.
For any specific $\omega$ we can then maximize Eq.~(\ref{eq:joint_Jensen}) with respect to $Q$ and $R$ to estimate $\ln \mathcal{Z}(\omega)$.
Better, we can allow $\omega$ to be another parameter in the model and thus optimize everything simultaneously.

The Karate Club graph \cite{zachary1977information} represents measured friendships among members of a university karate club.
At a later date, the club fractured into two different clubs.
The subsequent groups are often considered to represent a ground-truth on how the original graph should be divided.

\begin{figure}
  \centering
  \includegraphics[width=0.7\linewidth]{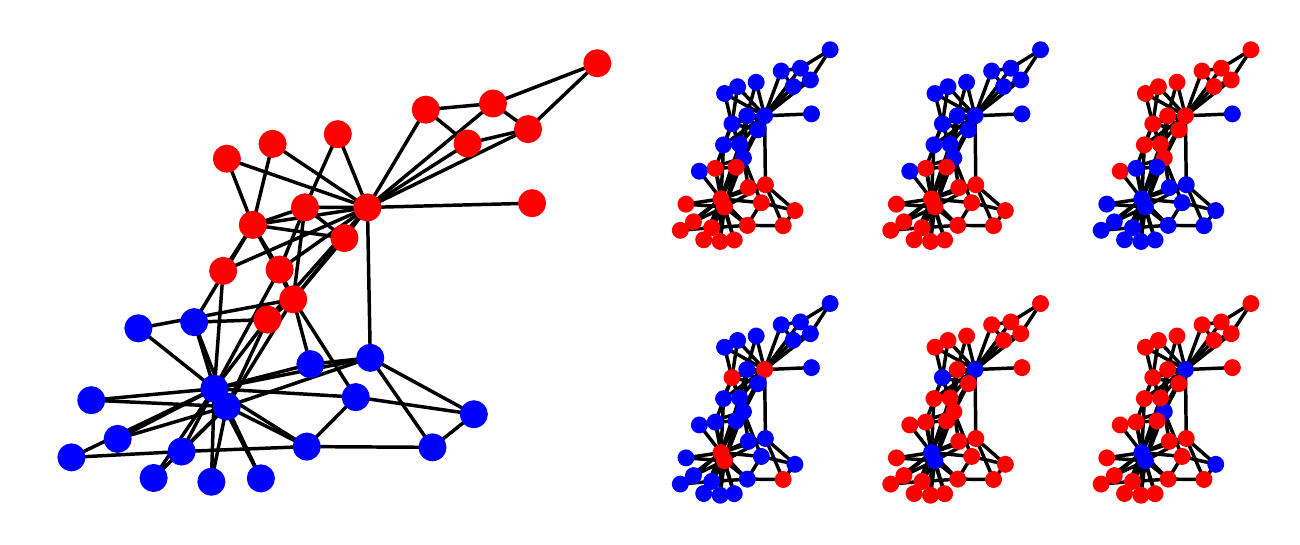}
  \caption{\label{fig:SBM} Karate club network.  Left: network colored to match the ``ground truth'' communities, i.e. the factions.  Right, top row: Samples from neural network at the factions local maximum. Right, bottom row: Samples from the neural network at the leaders-followers local maximum.
	}
\end{figure}

Our experiments find a local maximum for $\mathcal{Z}(\omega) = P(G \vert \omega)$ in which $\omega_{\mathrm{in}} > \omega_{\mathrm{out}}$.
We can sample $\bm{x}$ by first sampling $\bm{y}$ and then $\bm{x}$ from $R(\bm{x} \vert \bm{y})$.
This gives us independent samples of $\bm{x}$ that are (approximately) distributed according to the posterior distribution $P(\bm{x} \vert G, \omega)$.
We indeed find a distribution that is centered around the alleged ground-truth factions -- see Fig.~\ref{fig:SBM}.

However, our experiments also find a second local maximum, in which  $\omega_{\mathrm{in}} < \omega_{\mathrm{out}}$.
That is to say, a situation in which the network is split into two groups but where connections within each group are sparse and between are frequent.
This group also has an intuitive explanation: it is a leaders-followers dichotomy.
In fact, the maximum likelihood group assignment finds precisely this case \cite{pnas.0907096106, karrer2011stochastic}.

Thus we find two local optima: factions and leaders-followers. 
While the leaders-followers finds the single best maximum likelihood community division, it is not immediately obvious which optima is actually the preferred clustering of the network, according to the SBM.
In the factions posterior, both groups are roughly equally sized, whereas in the leaders-followers posterior, the groups are highly uneven.
Because there are exponentially more ways to split $n$ objects into roughly equally sized groups and highly uneven ones, it is possible that the factions grouping is preferred, i.e. that $P(G \vert \omega_{\mathrm{factions}}) > P(G \vert \omega_{\mathrm{leaders-followers}})$.
In actuality, we find that this is not the case -- the SBM truly favors the leaders-followers groupings.
Depending on what you are hoping to find by graph clustering, this provides a compelling reason to use the degree-corrected SBM \cite{karrer2011stochastic}.

Again, we should contrast our method with existing approaches for analyzing the SBM.
One approach is to use heuristics, such as the eigenvectors of an appropriately defined matrix  or belief propagation \cite{amini2013pseudo, krzakala2013spectral, decelle2011inference}.
These methods are generally fast but potentially highly biased.

An alternative approach for the SBM is again to use Markov chain Monte Carlo algorithms \cite{nowicki2001estimation}.
For example, one can use an Expectation Maximization (EM) algorithm in which the E-step is estimated by simulating an appropriate Markov chain.
Of course, this relies on constructing a high quality sampler, and it is also generally non-trivial to estimate the model evidence using this approach -- deciding which local maximum is preferred remains challenging.

\subsection{Ranking and noisy sorting}

For a final example, we consider the problem of ranking from noisy comparisons.
For a finite set of $n$ objects, we are given noisy comparisons that indicate which of two objects 
is ranked higher. We want to infer the underlying ranks.
Sports or chess matches provide prototypical examples of this scenario.

In the context of ranking, a state is a permutation of the $n$ objects.
We can represent this by letting $x_i \in \{ {1\over n}, {2\over n}, \dots {n\over n}\}$, where $x_i = k/n$ means object $i$ is ranked $k$th out of $n$.
We will assume each comparison obeys the true ordering with probability $w$, or points in the wrong direction with probability $1-w$.

If $i$ and $j$ were compared and $i$ beat $j$, we write $i \prec j$.
Any putative ranking either agrees or disagrees with the result of each comparison.
Let $V(\bm{x})$ be the total number of disagreements between ranking a $\bm{x}$ and the comparisons,
\begin{equation}
	V(\bm{x}) = \sum_{ i \prec j } \mathbb{I}( x_i > x_j )
	\label{eq:num_vios}
\end{equation}
with $\mathbb{I}$ being the indicator function.
If there are $m$ total comparisons then the posterior distribution over rankings is proportional to
\begin{equation}
	f(\bm{x}) = w^{m} \Big(\frac{1 - w}{w} \Big)^{V(\bm{x})}.
\end{equation}
We will apply the auxiliary variable machinery to estimate $\mathcal{Z} = \sum_{\bm{x}} f(\bm{x})$ and the distribution $P(\bm{x})=f(\bm{x}) / \mathcal{Z}$.
In general, exact calculation of $\ln \mathcal{Z}$ is \#P-complete; for $w=1$ it corresponds to counting linear extensions \cite{brightwell_counting_1991}.  Even finding a single best ranking is NP-complete; it is equivalent to the minimum feedback arc set problem.

In this example $x_i$ is discrete but can take any of $n$ different values and we have the constraint that $x_i \neq x_j$ for all $i,j$ (we don't allow for ties).
Equation~(\ref{eq:conditional_product_R}) and the Gumbel softmax trick 
no longer apply and so we need a new functional form for $R(\bm{x} \vert \bm{y})$ and a continuous relaxation to allow differentiation through the sampling procedure during training.

Thankfully, such a representation is available.
The $n$ dimensional unit hypercube can be split into $n!$ regular simplices, each defined as the set of points with a given sort-order.
For example, one such simplex is $\{\bm{z} : 0 < z_1 < z_2 < \dots z_n < 1 \}$.
Each vector of rankings $\bm{x}$ is at the center of one of these simplices and by definition Eq.~(\ref{eq:num_vios}) is identical at each point within a simplex. 
The regions where $x_i = x_j$ in the have zero volume, so by relaxing from $x_i \in \{{1\over n},\dots,{n\over n}\}$ to $x_i \in [0,1]$, we will obey $x_i \neq x_j$ but leave $\mathcal{Z} = \sum_{\bm{x}} f(\bm{x}) = n! \int_{[0,1]^n}f(\bm{x})\mathrm{d}\bm{x}$ invariant.\footnote{The factor of $n!$ is picked up because each simplex has volume $1/n!$.}

We then assume $R$ is a product of Beta distributions
\begin{equation}
	R(\bm{x} \vert \bm{y}) = \prod_i \frac{x_i^{\alpha_i(\bm{y})-1}(1-x_i)^{\beta_i(\bm{y})-1}} {B(\alpha_i(\bm{y}),\beta_i(\bm{y}))}
\end{equation}
and $\bm{\alpha}(\bm{y})$ and $\bm{\beta}(\bm{y})$ take the place of $\bm{\phi}(\bm{y})$ and are represented by a neural network.
Following the same logic of the Gumbel softmax trick, and to ensure pipeline is differentiable, we replace the non-differentiable indicator function in $V(\bm{x})$ with a steep sigmoid function, and reparametrize uniform noise using inverse transform sampling to generate $\bm{x}$.

\begin{figure}
  \centering
  \includegraphics[width=0.75\linewidth]{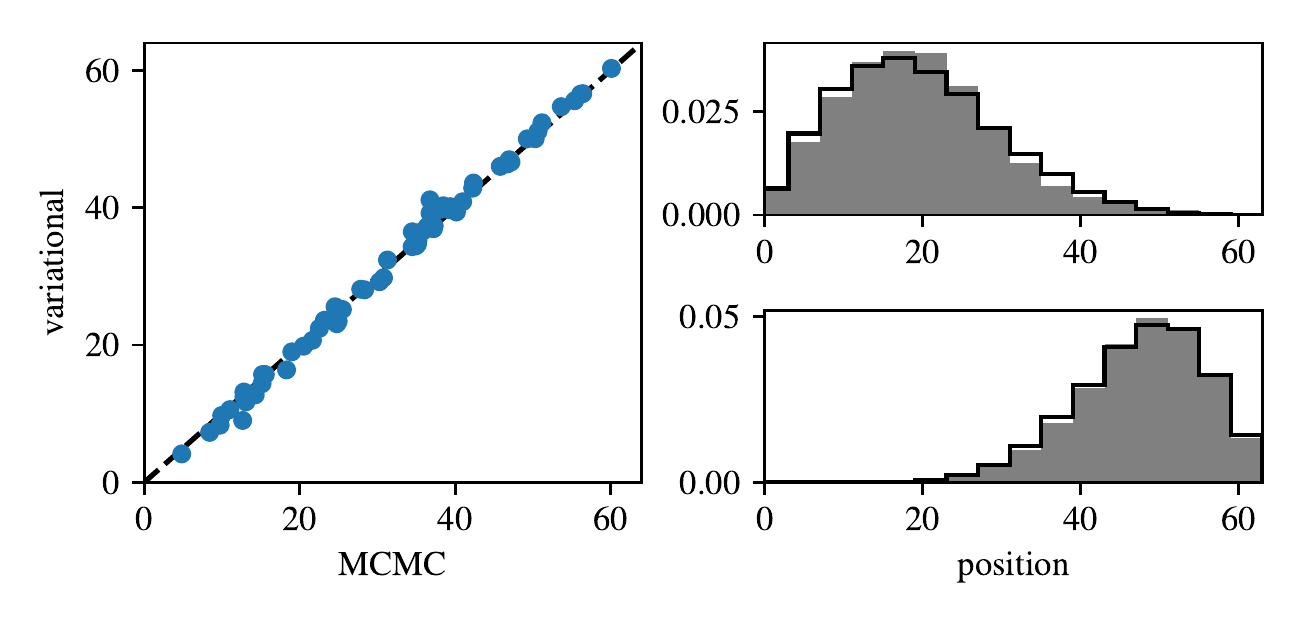}
  \caption{\label{fig:order} Distribution over rankings of $64$ objects from $512$ noisy comparison.  Left: The average location of objects, as estimated using Markov chain Monte Carlo and the variational approximations of this paper.  Right: The full distributions of position for two different objects.  In gray, we show a histogram of the distributions as estimated from MCMC.  The black line shows the inferred distribution from the variational method of this paper. }
\end{figure}

For the ranking experiments, we used synthetic data.
We fixed an arbitrary permutation of $64$ objects and randomly made $512$ noisy comparisons with noise parameter $w=0.75$.
We used this to define a distribution over permutations $P(\bm{x} \vert w)$.

We used the two neural networks, each with a single hidden layer with $1024$ units, using SELU activation functions.
The noise parameter $w$ could inferred by maximizing our estimate of $\ln \mathcal{Z}(w)$, however, we set it to the ground-truth value, $w=0.75$ so that we could compare against MCMC, which does not infer the parameters (although in principle an EM style algorithm could be used to achieve this).
Sample rankings were produced by first generating $\bm{y}$ from $P(\bm{y})$ and then $\bm{x}$ from $R(\bm{x}\vert\bm{y})$. Ranks were computed by finding which simplex the relaxed $\bm{x}$ fell inside, i.e. by the sort-order of its components.

We lack analytic values for $\ln \mathcal{Z}$ to compare against.
Instead, in Fig.~\ref{fig:order} we show comparisons to Markov chain Monte Carlo sampling for instances of $\bm{x}$; we find excellent agreement.

\section{Discussion}

In this paper we have considered variational methods for solving physics and statistical inference problems.
To this end we have repurposed the machinery of variational autoencoders but we have inverted the logic.
Instead of encoding data samples to random noise (such as a normal distribution), we attempt to ``encode'' random binary noise into the distributions of interest.

For the three tasks we consider -- Ising model, graph clustering, and ranking -- we find good performance using small and simple networks: a single hidden layer with $1024$ units.
As a result training takes seconds, even without dedicated hardware\footnote{Our code is available at \url{https://github.com/gcant/DistVAE}.}.
Presumably larger networks, and particularly those designed to match the structure of the problems, would improve performance.
For example convolutional neural networks would match the locality and translational invariance of the Ising model, or graph neural networks would match the structure of the graph clustering problem.

One benefit of our methods are their contiguity with traditional mean-field methods, due to the fact that we can vary the number of auxiliary variables.
For the Ising model, for example, at large $\beta$ (low temperature) a highly accurate approximation is found with only a single binary auxiliary variable.
The value of this binary variable switches between two solutions (majority of spins up vs. majority of spins down).
Likewise, for very small $\beta$ we are also able to find accurate approximations using only a few auxiliary variables.
Conversely, near the phase transition ($\beta = -\ln(\sqrt{2}-1)/2 \approx 0.4407$) accurate approximation requires using a larger number of auxiliary variables.

The examples we studied were chosen because they had known solutions – either exactly or approximately by using MCMC.
This allowed us to \emph{test} the VAE architecture, but in general it would be difficult to know whether the VAE has been successful.
Nevertheless, we can always guarantee that the approximation for $\mathcal{Z}$ is a lower bound.
For some problems this may be useful even if we cannot be sure how close the bound is to the true answer.

Future work should systematically explore the results of differing neural network architectures and differing numbers of auxiliary variables.
It should also address our lack of understanding of which distributions can and cannot be accurately represented by these feedforward neural networks.  
For example, these methods may be inappropriate for so-called \emph{glassy} systems, where both variational methods and MCMC can fail \cite{mezard1987spin}.
For extremely rich and complex distributions, i.e. those containing a very large number of roughly equal modes that are widely separated, the VAE may fail to find a good approximation.
If such a failure occurs, is it due to the training scheme, or a failure of the architecture itself?
At least in the case where feed-forward networks are used with a single hidden layer, as we considered in our experiments, it seems hopeful that analytic possibility/impossibility results may be achievable.

\acksection
This work was supported by NSF Grant BIGDATA-1838251.  I thank Cristopher Moore and Artemy Kolchinsky for helpful conversations.


\end{document}